\documentclass{article}
\usepackage{arxiv}
\usepackage[numbers]{natbib} 

\title{Transformer tricks: Precomputing the first layer}

\author{Nils Graef \\ \href{https://openmachine.ai}{OpenMachine},
  South San Francisco, CA 94080, \texttt{info@openmachine.ai}}

\begin{document} \maketitle

\begin{abstract}
This micro-paper \cite{micro-paper} describes a trick to speed up inference of transformers with RoPE \citep{RoPE} (such as LLaMA, Mistral, PaLM, and Gemma \citep{gemma}). For these models, a large portion of the first transformer layer can be precomputed, which results in slightly lower latency and lower cost-per-token.
Because this trick optimizes only one layer, the relative savings depend on the total number of layers. For example, the maximum savings for a model with only 4 layers (such as Whisper tiny \citep{Whisper}) is limited to 25\%, while a 32-layer model is limited to 3\% savings. See \citep{tricks} for code and more transformer tricks. \\
The next two sections detail the precompute for transformers with parallel attention/FFN \citep{parallel} (such as GPT-J, Pythia, and PaLM \citep{parallel, Pythia, PaLM}) and without (such as Llama 2, Mistral, and Mixtral \citep{LLaMA, Llama2, Mistral, Mixtral}).
\end{abstract}

\section{Precompute for parallel transformers}
\begin{figure}[h!] \centering  
  \includegraphics[scale=0.86]{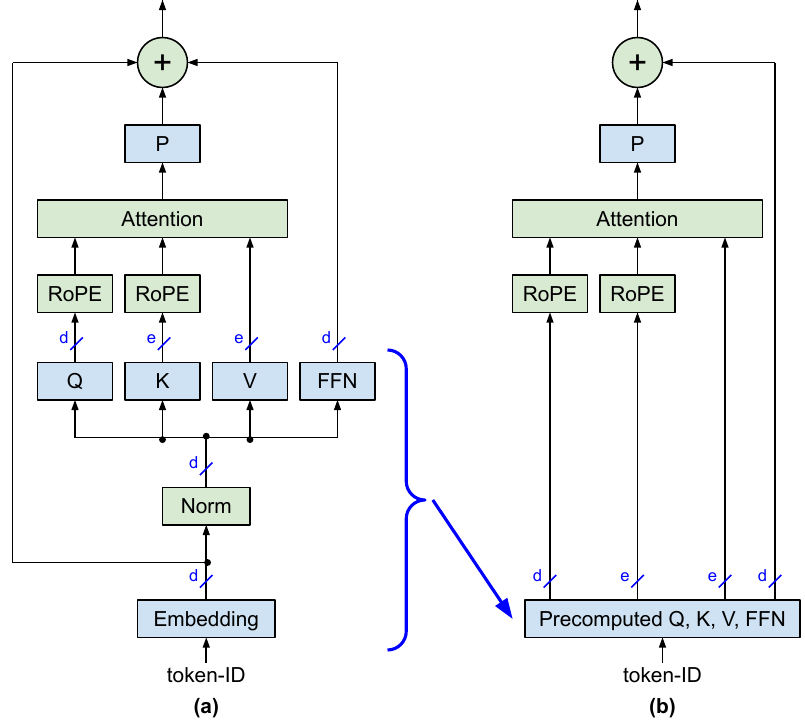}
  \caption{First layer of parallel transformer (a) without precompute; and (b) with precompute of FFN and linear layers Q, K, and V.}
 \label{fig1} \end{figure}

Figure \ref{fig1}(a) shows the first layer of a transformer with RoPE and parallel attention/FFN. Because the inputs of Q, K, V, and FFN only depend on the embedding, we can precompute their outputs and store them in memory instead of the input embeddings, see Figure \ref{fig1}(b).

Figure \ref{fig1} uses the following dimensions, based on the type of attention such as multi-head attention (MHA) \citep{vanilla}, multi-query attention (MQA) \citep{MQA}, and grouped-query attention (GQA) \citep{GQA}:

\begin{itemize}[topsep=-1pt]
  \item $d$: embedding dimension.
  \item $e$: $e = d$ for MHA. For MQA, $e = d / n_{heads}$. And for GQA, $e = d \cdot n_{kv\_heads} / n_{heads}$.
  \item Q, K, V, P are the linear layers for query, keys, values, and post-attention projection.
  \item FFN (feedforward network) is usually a two-layer MLP (multi-layer perceptron). Mistral and Llama2 use a two-layer MLP with a GLU variant \citep{GLU} for the first layer. And MoE models (mixture-of-experts) \citep{MoE} such as Mixtral use a switch FFN.
  \item The embedding layer is implemented by a simple memory read operation, where the token-ID provides the read-address to read $d$ values from memory.
\end{itemize}

The precompute is done as follows: For each token stored in the embedding table, perform the calculations needed for the first layer normalization, FFN, skip-connection, and linear layers Q, K, V, and store the results in memory instead of the original input-embeddings. This precompute is done offline only once and stored in the parameter memory (along with weights, biases, and output-embeddings).

The benefits of precompute include:
\begin{itemize}[topsep=-1pt]
  \item \textbf{Lower computational complexity per token}: For each token, we save the operations needed for FFN and the linear layers Q, K, V. This can speed up inference if the system is limited by compute.
  \item \textbf{Fewer memory reads for low batch sizes}: This can speed up inference for systems that are memory bandwidth limited, especially during the autoregressive next-token-prediction phase, see the table below and section \ref{sec:examples} for examples.
\end{itemize}

\begingroup \renewcommand{\arraystretch}{1.3} 
\begin{center} \begin{tabular}{r|l|l}
  & \textbf{Without precompute}            & \textbf{With precompute}  \\ \hline
    & \makecell[l]{1) For each token, read $d$ embedding values \\
      2) Plus, for each batch, read weights for Q, K, V, FFN}
    & \makecell[l]{For each token, read $2(d+e)$ \\  precomputed values} \\ \hline
  \makecell[l]{Reads per batch: \\ ($B$ is batch-size)} & $B \cdot d + \verb+num_weights_Q_K_V_FFN+$ & $B \cdot 2(d+e)$
\end{tabular} \end{center} \endgroup


Notes on batch size:
\begin{itemize}[topsep=-1pt]
  \item During the prefill phase, many implementations use a batch size larger than 1, because the input tokens can be processed in parallel.
  \item During the autoregressive next-token-generation phase, single-user implementations often use a batch size of \verb+num_beams+ (i.e. the width of the beam search, such as \verb+num_beams+ = 4), while multi-user implementations use larger batch sizes. However, the maximum batch size for multi-user applications can be limited by the total memory capacity as the number of KV-caches increases linearly with the batch size.
\end{itemize}

However, precomputing the first layer can increase (or decrease) the total memory size, which depends on the vocabulary size and the number of eliminated weights as shown in the table below. For example, the total memory size of Mistral-7B only increases by 2\%, see section \ref{sec:examples} for more details.

\begingroup \renewcommand{\arraystretch}{1.3} 
\begin{center} \begin{tabular}{l|l}
  \textbf{Without precompute}                      & \textbf{With precompute} \\ \hline
  1) Store embeddings: $d \cdot \verb+vocab_size+$ & Store precomputed values: $2(d+e) \cdot \verb+vocab_size+$ \\
  2) Store weights for Q, K, V, and FFN            &                          \\ \hline
\end{tabular} \end{center} \endgroup

\section{Precompute for serial transformers}
\begin{figure} \centering
  \includegraphics[scale=0.86]{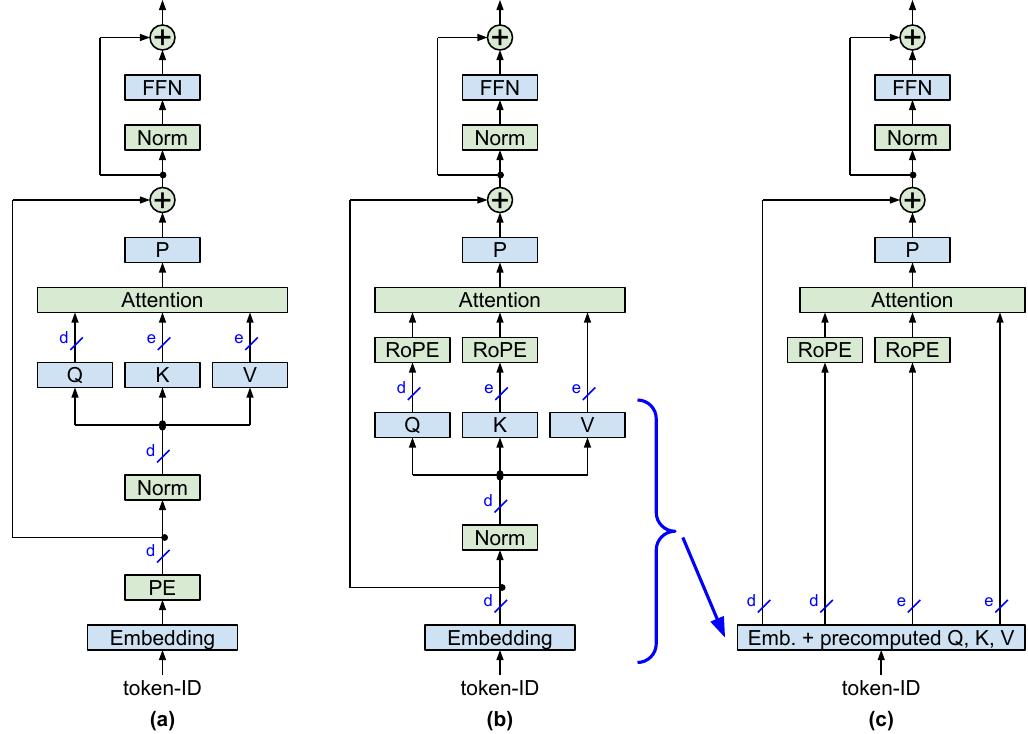}
  \caption{First transformer layer. (a) Vanilla with pre-normalization and vanilla PE; (b) Vanilla with RoPE; (c) Precomputing linear layers Q, K, V.}
 \label{fig2} \end{figure}

Transformers without the parallel attention/FFN scheme can also benefit from precompute, but the savings are smaller: As shown in Figure \ref{fig2}(c), we can only precompute Q, K, and V, but not the FFN. For reference, Figure \ref{fig2}(a) shows the vanilla transformer with absolute positional encoding (PE) instead of RoPE and with pre-normalization \citep{pre-norm}. The PE is located right after the embedding layer, which prevents us from precomputing the first layer. But replacing the PE by RoPE, as done in Figure \ref{fig2}(b), allows us to precompute the linear layers Q, K, and V and store the precomputed values along the embeddings in memory as illustrated in Figure \ref{fig2}(c).

\section{Examples} \label{sec:examples}

\begingroup
\renewcommand{\arraystretch}{1.3} 
\begin{center} \begin{tabular}{|l|c|c|c|l|} \hline
  \textbf{Parameter} & \textbf{Pythia-6.9B} & \textbf{Mistral-7B} & \textbf{Mixtral-8x7B} & \textbf{Notes} \\ \hline
  Parallel attention/FFN? & parallel         & \multicolumn{2}{c|}{serial} & \citep{parallel}              \\ \hline
  MHA, MQA, or GQA?       & MHA              & \multicolumn{2}{c|}{GQA}    & \citep{vanilla, MQA, GQA}     \\ \hline
  \verb+dim+ (aka $d$)    & \multicolumn{3}{c|}{4,096}                     & embedding dimension           \\ \hline
  \verb+n_layers+         & \multicolumn{3}{c|}{32}                        & number of layers              \\ \hline
  \verb+n_heads+, \verb+n_kv_heads+ & 32, 32 & \multicolumn{2}{c|}{32, 8}  & number of heads, KV-heads     \\ \hline
  \verb+e+ (output dim. of K, V) & 4,096     & \multicolumn{2}{c|}{1,024}  & \verb+e = d * n_kv_heads / n_heads+ \\ \hline
  FFN type                & 2-layer MLP      & SwiGLU *) & SwiGLU MoE      & *) MLP with SwiGLU (GLU variant) \citep{GLU, MoE} \\ \hline
  FFN \verb+hidden_dim+   & 16,384           & \multicolumn{2}{c|}{14,336} & FFN hidden dimension          \\ \hline
  FFN \verb+n_experts+    & \multicolumn{2}{c|}{1}  & 8                    & FFN number of experts         \\ \hline
  \verb+vocab_size+       & 50,400           & \multicolumn{2}{c|}{32,000} & vocabulary size               \\ \hline

  \multicolumn{5}{|l|}{\textbf{Number of weights (calculated from above parameters):}}                                 \\ \hline
  Q+P weights per layer & \multicolumn{3}{c|}{33,554,432}                & \verb+2 * dim * dim+                        \\ \hline
  K+V weights per layer & 33,554,432  & \multicolumn{2}{c|}{8,388,608}   & \verb+2 * dim * dim / n_heads * n_kv_heads+ \\ \hline
  FFN weights per layer & 134,217,728 & 176,160,768 & 1,409,286,144      & \verb+(2 or 3) * dim * hidden_dim * n_exp.+ \\ \hline
  Input+output embed.   & 412,876,800 & \multicolumn{2}{c|}{262,144,000} & \verb+2 * dim * vocab_size+                 \\ \hline
  \multicolumn{1}{|r|}{\textbf{Total weights:}} & 6.9B & 7.2B &  46.7B   &                                             \\ \hline
\end{tabular} \end{center}
\endgroup

The table above compares the configurations and number of weights of Pythia-6.9B, Mistral-7B, and Mixtral-8x7B. The next table shows the memory read savings and memory size increases for Pythia-6.9B, Mistral-7B, and a hypothetical Mixtral-8x7B with parallel attention/FFN layers.

\begingroup
\renewcommand{\arraystretch}{1.4} 
\begin{center} \begin{tabular}{|l|c|c|>{\centering\arraybackslash}m{7.8em}|} \hline
  & \textbf{Pythia-6.9B} & \textbf{Mistral-7B} & \textbf{Hypothetical Mixtral-8x7B with parallel attn./FFN} \\ \hline
  Number of weights that can be eliminated    & 184,549,376 & 25,165,824 & 1,434,451,968 \\ \hline
  Number of reads w/o precompute for batch 1  & 184,553,472 & 25,169,920 & 1,434,456,064 \\ \hline
  Number of reads with precompute for batch 1 & 16,384      & 10,240     & 10,240      \\ \hline
  \multicolumn{1}{|r|}{\textbf{First layer reduction factor for batch size 1:}} & \textbf{11,264x} & \textbf{2,458x} & \textbf{140,084x} \\ \hline
  \multicolumn{1}{|r|}{\textbf{First layer reduction factor for batch size 16:}}    & 704x    & 154x   & 8,756x     \\ \hline
  \multicolumn{1}{|r|}{\textbf{First layer reduction factor for batch size 256:}}   & 44x     & 10x    & 548x        \\ \hline
  \multicolumn{1}{|r|}{\textbf{First layer reduction factor for batch size 1,024:}} & 11x     & 3x     & 137x         \\ \hline

  \multicolumn{4}{|l|}{\textbf{Increase (or decrease) of total weight memory size:}}                               \\ \hline
  Increase embedding memory by $(2e + d) \cdot \verb+vocab_size+$ & 619,315,200 & \multicolumn{2}{c|}{196,608,000} \\ \hline
  Memory decrease due to elimination of weights & –184,549,376 &  –25,165,824   & -1,434,451,968                    \\ \hline
  \multicolumn{1}{|r|}{\textbf{Total absolute memory increase (or decrease):}}  &  434,765,824 & 171,442,176 & \textbf{-1,237,843,968}  \\ \hline
  \multicolumn{1}{|r|}{\textbf{Total relative memory increase (or decrease):}}  & 6\%          & \textbf{2\%}         & \textbf{–3\%} \\ \hline
\end{tabular} \end{center}
\endgroup

\section*{Acknowledgements}
We would like to thank \href{https://scholar.google.com/citations?user=LlK_saMAAAAJ&hl=en}{James Martens (DeepMind)} for his generous support and endorsement for the arXiv submission process.

\bibliographystyle{unsrtnat}
\bibliography{references}

\end{document}